\documentclass[acmlarge,screen,nonacm]{acmart}
\AtBeginDocument{%
  \providecommand\BibTeX{{%
    Bib\TeX}}}

\usepackage{amsmath, amsthm}
\usepackage{subcaption}
\usepackage{geometry}

\copyrightyear{2025}
\acmYear{2025}
\setcopyright{rightsretained}
\acmConference[GECCO '25 Companion]{Genetic and Evolutionary Computation
Conference}{July 14--18, 2025}{Malaga, Spain}
\acmBooktitle{Genetic and Evolutionary Computation Conference (GECCO '25
Companion), July 14--18, 2025, Malaga,
Spain}\acmDOI{10.1145/3712255.3726643}
\acmISBN{979-8-4007-1464-1/2025/07}

\AtBeginDocument{%
  \providecommand\BibTeX{{%
    \normalfont B\kern-0.5em{\scshape i\kern-0.25em b}\kern-0.8em\TeX}}}

\begin{document}

\title{Comparing Optimization Algorithms Through the Lens of Search Behavior Analysis}
\author{Gjorgjina Cenikj}
\affiliation{
  \institution{Computer Systems Department\\ Jo\v{z}ef Stefan Institute\\ Jo\v{z}ef Stefan International Postgraduate School}
  \city{Ljubljana} 
  \country{Slovenia}
}
\email{gjorgjina.cenikj@ijs.si}

\author{Gašper Petelin}
\affiliation{
  \institution{Computer Systems Department\\ Jo\v{z}ef Stefan Institute\\ Jo\v{z}ef Stefan International Postgraduate School}
  \city{Ljubljana} 
  \country{Slovenia}
}
\email{gasper.petelin@ijs.si}

\author{Tome Eftimov}
\affiliation{
  \institution{Computer Systems Department\\ Jo\v{z}ef Stefan Institute}
  \city{Ljubljana} 
  \country{Slovenia}
}
\email{tome.eftimov@ijs.si}

\renewcommand{\shortauthors}{Cenikj et al.}

\begin{abstract}
The field of numerical optimization has recently seen a surge in the development of "novel" metaheuristic algorithms, inspired by metaphors derived from natural or human-made processes, which have been widely criticized for obscuring meaningful innovations and failing to distinguish themselves from existing approaches. Aiming to address these concerns, we investigate the applicability of statistical tests for comparing algorithms based on their search behavior. 
We utilize the cross-match statistical test to compare multivariate distributions and assess the solutions produced by 114 algorithms from the MEALPY library. These findings are incorporated into an empirical analysis aiming to identify algorithms with similar search behaviors.
\end{abstract}



\maketitle

\section{Introduction}
In the past few decades, the field of numerical optimization has experienced an influx of so-called "novel" metaheuristic methods, inspired by metaphors derived from natural or human-made processes~\cite{stork2022new}. From the behaviors of various animal species to the distribution of mathematical operations, and nuclear reaction processes, seemingly any conceivable concept could be used as a foundation for introducing new metaheuristics. As argued in~\cite{metaheuristics_metaphor_exposed} this trend poses a risk to the scientific rigor within the field of metaheuristics, and can often be "a step backward rather than forward", and "distracts attention away from truly innovative ideas in the field of metaheuristics". The gravity of this trend has led to a call-to-action to stop the publication of such "novel" algorithms, which has been signed by almost 100 researchers in the field of optimization~\cite{call_to_action}. Yet, the evaluation of an algorithm's novelty remains a challenge. As highlighted in~\cite{metaheuristics_metaphor_exposed}, one of the factors contributing to the difficulty in evaluating the novelty of a proposed algorithm is the fact that describing the algorithm using the terminology of the chosen metaheuristic obscures the fundamental algorithm behaviour~\cite{camacho_intelligent_water_drops}. Additionally, authors often fail to position the algorithm in the metaheuristics literature and define its relation to other algorithms.

\textbf{Our contribution:} We aim to explore an empirical approach to the comparison of algorithms that takes into account their search behavior by analyzing the solutions explored during the optimization process. We employ the cross-match statistical test~\cite{crossmatch_test} for comparing multivariate distributions of the solutions generated by 114 algorithms from the MEALPY library~\cite{mealpy} on the Black Box Optimization Benchmarking (BBOB)~\cite{bbob} suite. An empirical analysis is then performed to identify algorithms that exhibit similar search behaviors.

\textbf{Reproducibility:} The code for conducting the experiments is publicly available at https://github.com/gjorgjinac/optimization\_
algorithm\_statistical\_comparison.

\section{Background: Crossmatch Test for Comparing Multivariate Distributions}

The crossmatch test~\cite{crossmatch_test} is a nonparametric, distribution-free statistical method for comparing two multivariate distributions based on the adjacency relationships among observations in a combined dataset.
Let \( X = \{x_1, x_2, \ldots, x_m\} \) and \( Y = \{y_1, y_2, \ldots, y_n\} \) represent two independent samples of size \( m \) and \( n \), drawn from distributions \( F_X \) and \( F_Y \), respectively. The goal of a test for comparing two multivariate distributions is to evaluate the null hypothesis: $H_0: F_X = F_Y$
against the alternative hypothesis: $H_1: F_X \neq F_Y.$
The crossmatch test includes the following steps:

   \textbf{Adjacency Graph Construction:} 
    The samples of points from both distributions, $X$ and $Y$, are combined into a single set of size \( m + n \). The distances between the points from the combined set are used to divide the \( m + n \) points into pairs, in such a way that the total distance within pairs is minimized. A crossmatch occurs if a point from \( X \) is paired with a point from \( Y \). The total number of crossmatches is denoted by \( C \).

     \textbf{Test Statistic Calculation:} 
    The observed number of crossmatches, \( C \), is used as the test statistic. Under the null hypothesis, the expected number of crossmatches is determined based on random assignment of labels to the combined data.

   \textbf{P-value Computation:} 
    A p-value is computed by comparing the observed number of crossmatches, \( C \), to its null distribution. Small p-values indicate significant differences between the distributions.

\section{Methodology}
\label{sec:methodology}
Our approach involves the comparison of candidate solutions explored by the algorithms during the optimization process, with the steps defined in continuation.

 \textbf{Algorithm Execution: } All algorithms to be compared are executed on the same suite of optimization problem instances several times, with fixed random seeds in such a way that initial populations of the algorithms are shared under the same random seed.
 
    \textbf{Scaling: } A min-max scaling of the populations explored by all algorithms is performed by merging the trajectories from all executions of all algorithms for a single problem instance, and scaling both the objective function values and the candidate solutions from all trajectories for the same problem instance. This ensures that the scaled populations of the same problem instance, but different algorithms, are comparable. 
    
 \textbf{Statistical Testing: } We use the implementation of the crossmatch test~\cite{crossmatch_test} in the \textit{crossmatch} R package to test whether the populations produced by two algorithms on a single problem instance come from the same distribution. In particular, given algorithms $a_1$ and $a_2$ executed on an optimization problem instance $o$ for $I$ iterations and $R$ runs (i.e., repetitions), we compare the populations generated by the two algorithms on a fixed problem instance, fixed iteration, and fixed run. 
    More precisely, denoting by $r$ the run number, $r \in \{1..R\}$ and by $i$ the iteration number, $i \in \{1..I\}$, we compare the population $p_{o,a_1,i,r}$ produced by algorithm $a_1$ in run $r$ and iteration $i$ on problem instance $o$ to the population $p_{o,a_2,i,r}$ produced by algorithm $a_2$ on the same problem instance, in the same iteration, and run.
    We execute the test on each pair of populations, with the same p-value of 0.05, applying the Bonferroni correction~\cite{weisstein2004bonferroni} for multiple comparisons within the same run. It is important to note that each pairwise comparison involves two independent samples, representing the populations generated by two different algorithms at the same iteration of the run. Additionally, we emphasize that we are not comparing populations across different iterations of the run. The distance between individuals is captured using the euclidean distance.
    This yields a statistical outcome for the populations at each iteration of the search process of two algorithms executed on the same problem with the same random seed ($o,a_1,a_2,i,r$). 
    
   \textbf{Empirical aggregation of outcomes:} To measure the similarity between a pair of algorithms, we use the statistical test results to define an empirical heuristic. Specifically, we calculate the percentage (ratio) of iterations from a run on a given problem for which the statistical test fails to reject the null hypothesis, indicating that the two populations likely originate from the same distribution. For each pair of algorithms, we aggregate this percentage by calculating its mean value across all problems and runs to derive a similarity indicator.

\section{Experimental Design}
First, we describe the selected benchmark suite of problem instances used in our analysis, followed by the portfolio of algorithms evaluated on this benchmark suite.

\textbf{Benchmark suite:}
As a benchmark suite, BBOB~\cite{bbob} is used. The benchmark contains 24 problem classes, each with multiple instances and different dimensions. We use the first instance of each problem with problem dimension $d \in {\{2, 5\}}$. 

\textbf{Optimization algorithm portfolio:}
\sloppy{The MEALPY library~\cite{mealpy} is an open-source Python library for metaheuristic optimization with a diverse selection of algorithms.} 
\sloppy{The performance data has been utilized from a previous study~\cite{dv_largescale_benchmarking} and has been collected using the IOHExperimenter platform~\cite{ioh_experimenter}.
Each algorithm is executed on each problem instance five times with a different random seed and a budget of 500$d$ function evaluations. The population size is set to 50 for all algorithms.
We compare the algorithms by running statistical analysis on the trajectories of the algorithms executed on the same problem instance with the same random seed. We remove algorithms for which the initial population does not match the initial population of the remaining algorithms under the same random seed. This results in a total of 114 algorithms to be compared with the following distribution within the eight MEALPY groups: \textit{bio\_based}	(11), \textit{evolutionary\_based} (11), \textit{human\_based} (21), \textit{math\_based} (7), \textit{music\_based} (2), \textit{physics\_based} (11), \textit{swarm\_based} (44), \textit{system\_based} (7).}

\section{Results}
\begin{figure}
\centering
\begin{subfigure}[b]{0.4\linewidth}
\centering
\includegraphics[width=\linewidth]{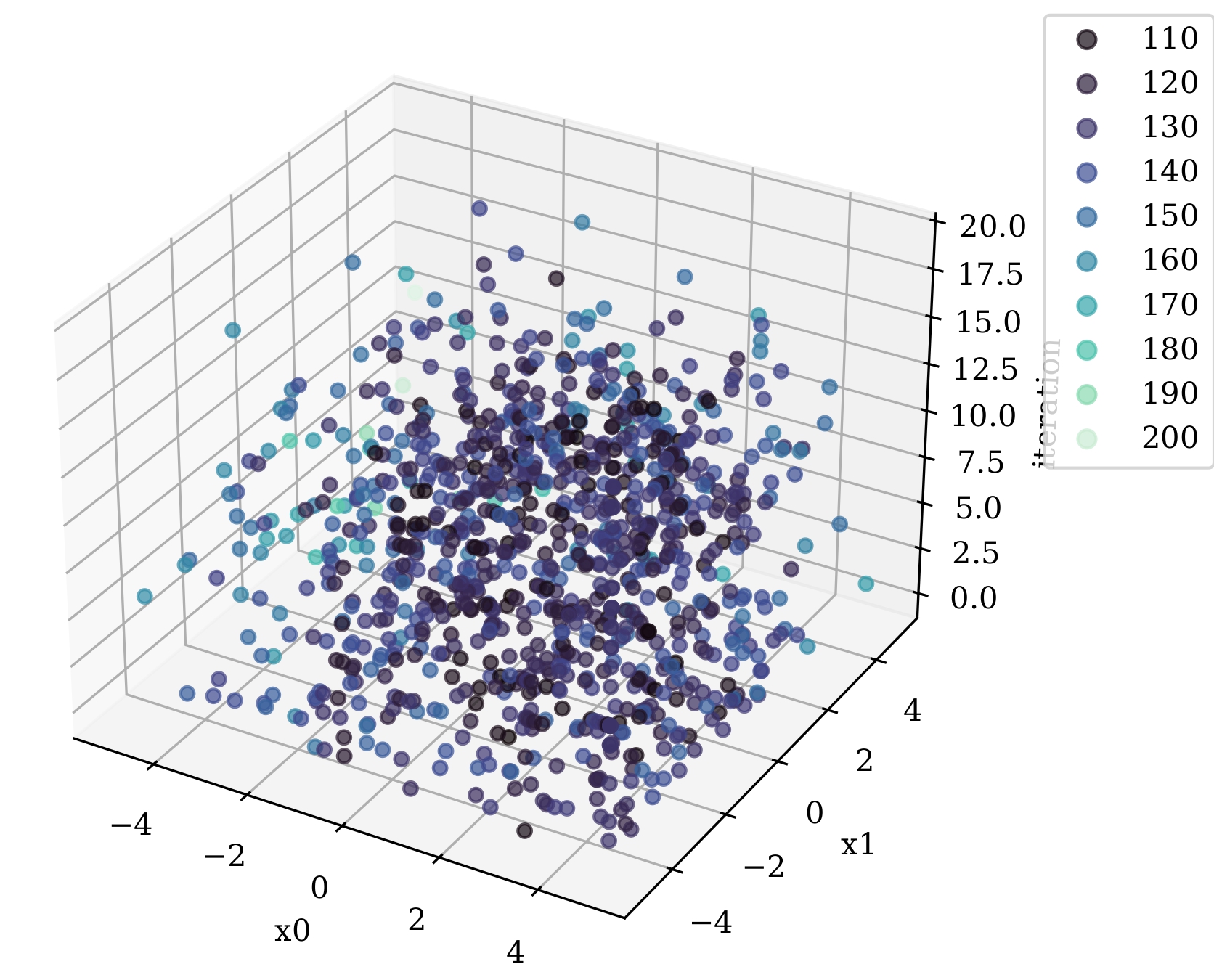}
\caption{BaseDE}
\label{fig:baseDE}
\end{subfigure}
\hfill
\begin{subfigure}[b]{0.4\linewidth}
\centering
\includegraphics[width=\linewidth]{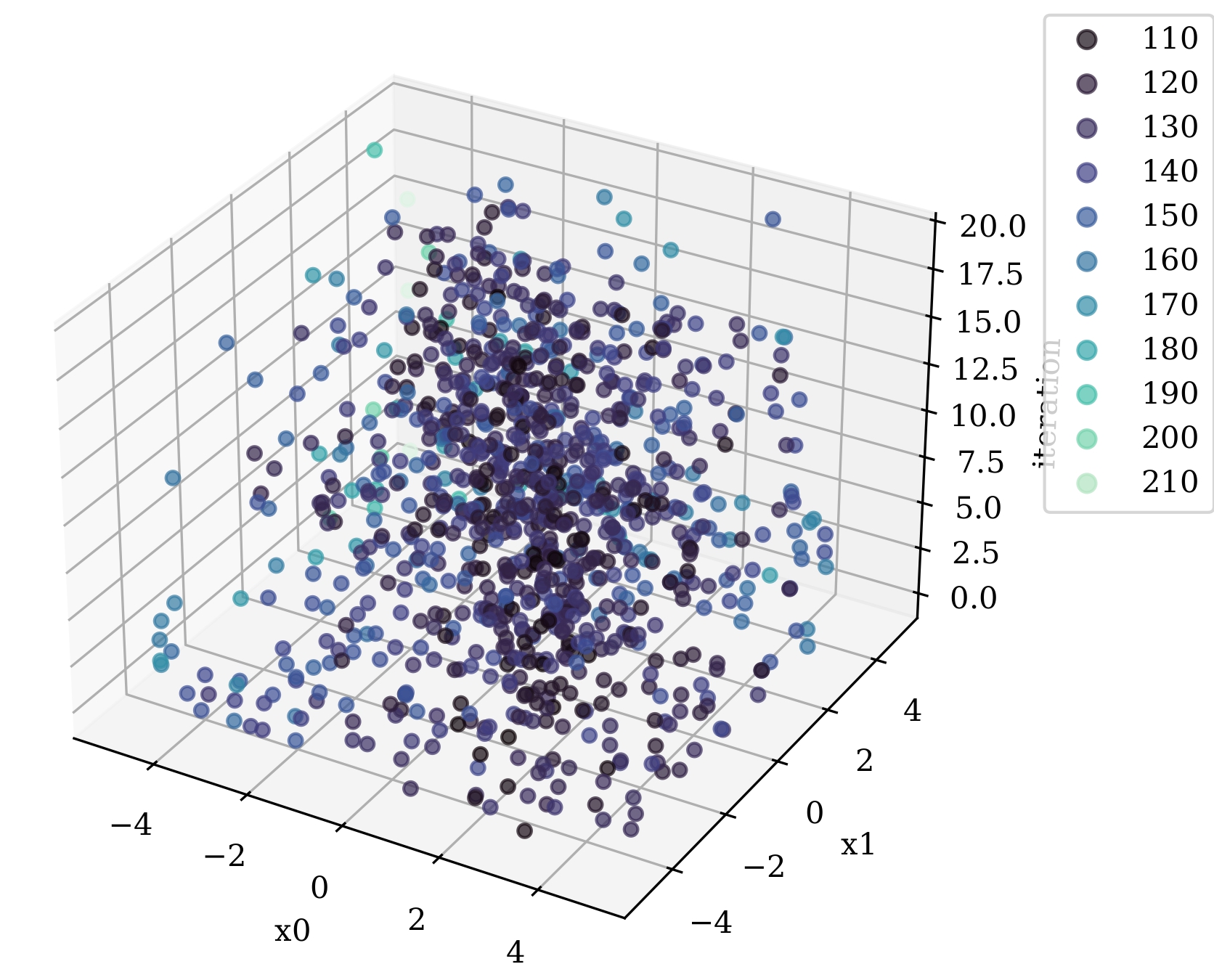}
\caption{SADE}
\label{fig:SADE}
\end{subfigure}

\vspace{10pt}

\centering
\begin{subfigure}[b]{0.4\linewidth}
\centering
\includegraphics[width=\linewidth]{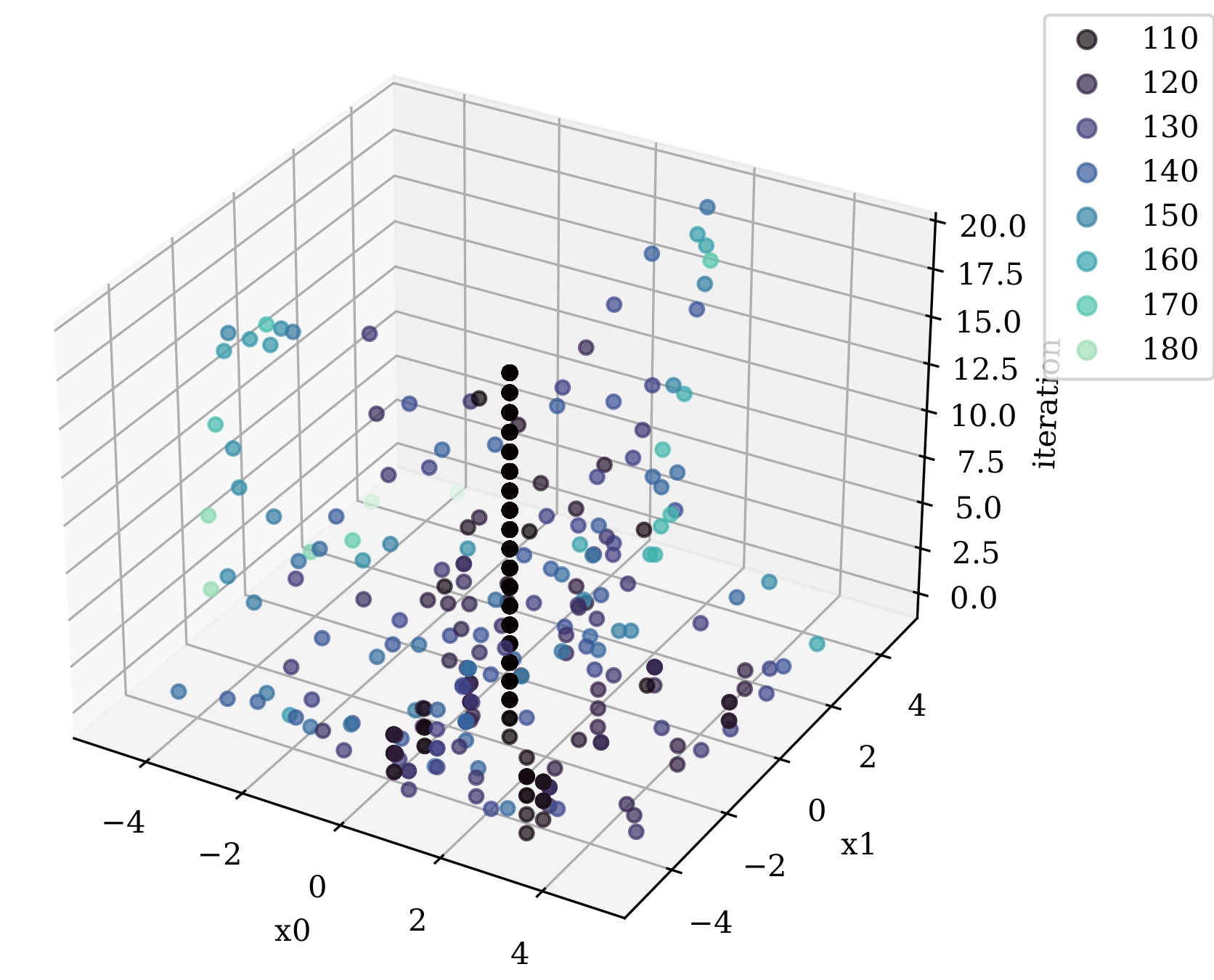}
\caption{BaseGA}
\label{fig:baseGA}
\end{subfigure}

\caption{Scatterplot of the candidate solutions explored by each of the three algorithms: BaseDE, SADE, BaseGA in one execution on the first instance of the 24th 2$d$ BBOB problem class}
\label{fig:scatterplot_trajectory_example}
\end{figure}

To demonstrate the applicability of the statistical test in our use case, we start by showing how the test behaves when applied on a single pair of trajectories. As an example, we compare the trajectories of the BaseDE (Differential Evolution) algorithm to the SADE (Self-Adaptive Differential Evolution) and BaseGA (Genetic Algorithm) algorithms, executed on the first instance of the 24th 2$d$ BBOB problem class. We present the example on 2$d$ problems, since they are most straightforward to visualize without a loss of information which would be induced by dimensionality reduction. 
The algorithms are initialized with the same population and are executed for 20 iterations. Figure~\ref{fig:scatterplot_trajectory_example} presents the candidate solutions explored by each of the algorithms. The visualization is generated following examples in~\cite{tea_visualizing}. Each subplot contains the trajectory of a different algorithm. The axes on the bottom, with range [-5,5], represent the values of the candidate solutions in the two dimensions. The vertical axis, with range [0,20], represents the iteration number, while the color indicates the objective function value of the candidate solution. It can be observed that the trajectory of the BaseGA algorithm is very different from the trajectories of the BaseDE and SADE algorithms. 

\begin{figure}

\centering
\includegraphics[width=0.95\linewidth]{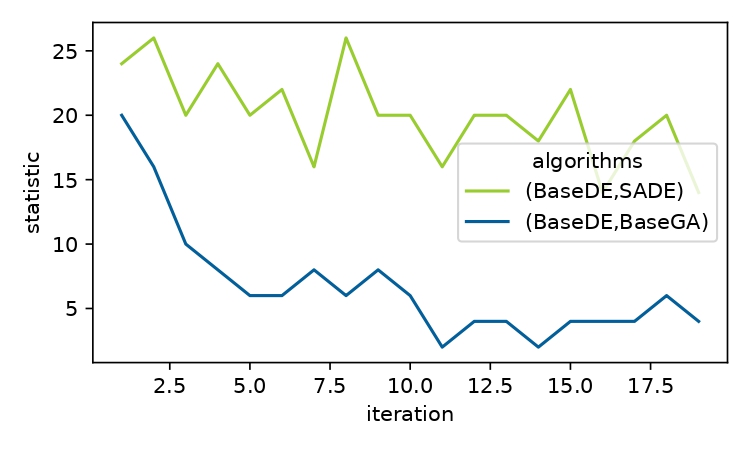}

\label{fig:crossmatch_statistic_example}

\caption{Values of the test statistic obtained with both tests for the algorithms (BaseDE, SADE) and (BaseDE, BaseGA) executed on the first instance of the 24th 2$d$ BBOB problem class}
\label{fig:crossmatch_statistic_example}
\end{figure}

Figure~\ref{fig:crossmatch_statistic_example} shows the value of the crossmatch statistic throughout the different iterations of the search process. Please note that the maximal value the statistic can obtain is 50, since this is the population size set for all algorithms, and we are comparing the algorithms at a population level. The differently colored lines depict the two different pairs of algorithms being compared, i.e., (BaseDE, SADE) and (BaseDE, BaseGA). Looking at the value of the crossmatch statistic in Figure~\ref{fig:crossmatch_statistic_example}, we can see a decreasing trend for the (BaseDE, BaseGA) algorithms (represented by the blue line). The crossmatch statistic has a value of 20 in the first iteration, indicating that the populations of both algorithms are similar, which is expected, since they start from the same initial population. 
As the search process continues, the value of the statistic drops, meaning that the populations explored are not as similar. On the other hand, looking at the green line representing the (BaseDE, SADE) algorithm pair, we can see that the line is somewhat oscilating around the value of 20, however, it remains mostly above the value of 15 throughout the entire search process. This means that on this problem, the SADE algorithm is much more similar to BaseDE than the BaseGA algorithm, which makes sense, since SADE is a variant of the DE algorithm. 

\begin{figure}
    \centering
    \includegraphics[width=0.4\linewidth]{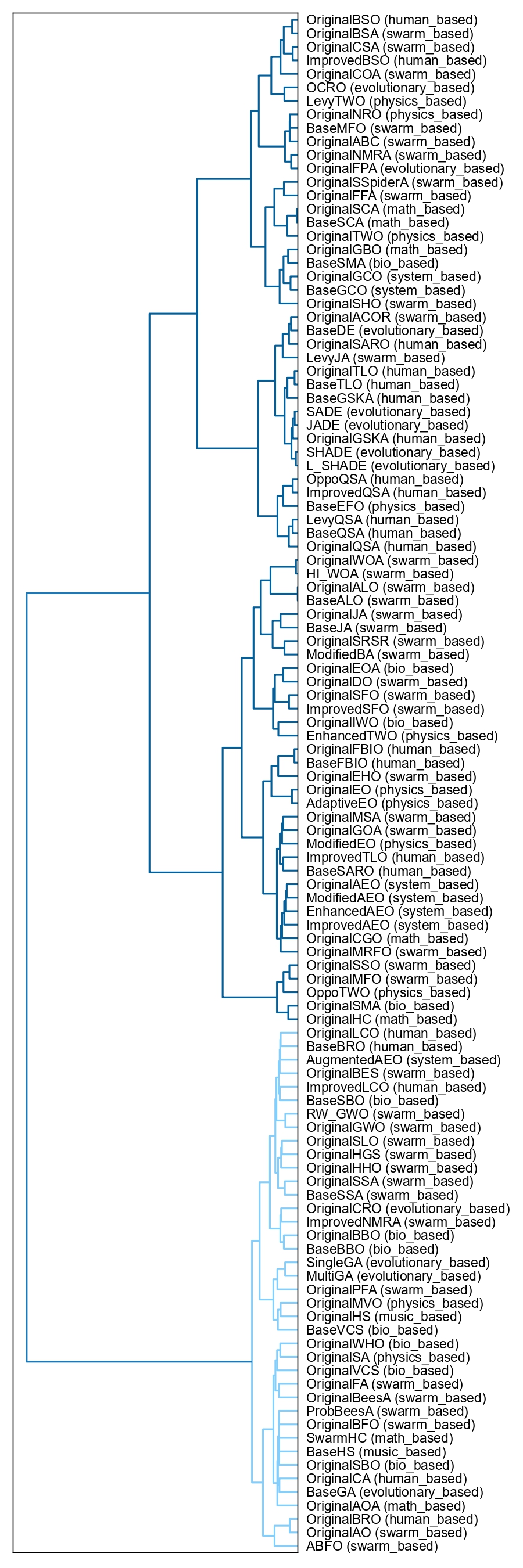}
    \caption{Algorithms grouped in a dendrogram structure}
    \Description{Algorithms grouped in a dendrogram structure}
    \label{fig:dendrogram}
\end{figure}

\sloppy{Next, we analyze the obtained algorithm similarities, defined by the empirical heuristic (i.e., the percentage (ratio) of iterations from a run on a given problem for which the statistical test fails to reject the null hypothesis, aggregated across all problems and runs, in both problem dimensions). Figure~\ref{fig:dendrogram} depicts a hierarchical grouping of the algorithms based on the similarities in their search behaviour. The dendrogram is constructed by grouping algorithms using the Ward variance minimization algorithm~\cite{murtagh2014ward} where distance is captured by the mean score obtained across both problem dimensions.
We can observe several groupings of similar algorithm variants being formed: Physics-based Equilibrium Optimization algorithms (OriginalEO, AdaptiveEO); System-based Artificial Ecosystem Optimization  algorithms (ImprovedAEO, EnhancedAEO, OriginalAEO, ModifiedAEO);
Human-based Queuing Search Algorithm algorithms (BaseQSA, LevyQSA, OriginalQSA, OppoQSA, ImprovedQSA); Evolutionary-based Differential Evolution algorithms (SHADE, L\_SHADE, JADE, SADE);
Evolutionary-based Genetic Algorithm (GA) algorithms (SingleGA, MultiGA).}

Other algorithm pairs which are linked together are the implementations of the same algorithm (OriginalGCO, BaseGCO), (OriginalSCA, BaseSCA), (OriginalTLO, BaseTLO), (BaseJA, OriginalJA).
We can also observe many algorithms belonging to different MEALPY groups which are linked together, indicating that even though these algorithms derive inspiration from different processes, they exhibit similar search behavior.

Nevertheless, we would like to point out that some algorithm pairs exhibiting high similarity scores (low rates of rejecting the null hypothesis) may not have completely identical trajectories. 
Taking this into account, one can also look into the values of the test statistic when interpreting the results of the statistical testing. The test statistic can also be used as an additional indicator of similarity.

\section{Conclusion}
In this study, we introduced a novel empricial technique for comparing the search trajectories of optimization algorithms. Our findings revealed substantial similarities between certain algorithm pairs, particularly those where multiple variants are derived from the same base algorithm.
More interestingly, our approach also revealed similarities between seemingly unrelated algorithms based on different metaphors or belonging to entirely different groups, where further investigation is needed.
This approach provides a valuable metric for comparing newly proposed metaheuristic algorithms against existing ones, allowing researchers to assess the degree of similarity in their behavior and encourages the development of algorithms that bring meaningful improvements. 

\section*{Acknowledgments}
We acknowledge the support of the Slovenian Research and Innovation Agency through program grant No.P2-0098, young researcher grants No.PR-12393 to GC and No. PR-11263 to GP, and project grants No.J2-4460 and No. GC-0001. This work is also funded by the European Union under Grant Agreement No.101187010 (HE ERA Chair AutoLearn-SI) and the EU Horizon Europe program (grant No. 101077049, CONDUCTOR).

\end{document}